\documentclass[journal]{IEEEtran}

\ifCLASSINFOpdf
\else
\fi

\usepackage{bbm}
\usepackage{float}
\usepackage{hyperref}
\usepackage{algorithmicx}
\usepackage{mathtools}
\usepackage{authblk}
\usepackage{booktabs}
\usepackage{multirow}
\usepackage{adjustbox}
\usepackage{subcaption}
\usepackage[bottom]{footmisc}
\usepackage{xcolor}
\begin{document}

\title{Self-supervised Contrastive Learning for Volcanic Unrest Detection}

\author[12]{Bountos~Nikolaos~Ioannis\thanks{This  work  has  received  funding  from  the  European  Union’s  Horizon2020  research  and  innovation  project  DeepCube,  under  grant  agreement number 101004188.}}
%\thanks{Nikolaos Ioannis Bountos is with both the Institute of Astronomy, Astrophysics, Space Applications \& Remote Sensing, National Observatory of Athens (IAASARS-NOA), Vas. Pavlou & I. Metaxa, GR-15 236 Penteli, Greece, and the Department of Informatics \& Telematics of Harokopio University of Athens (IT-HUA), El. Venizelou 70, Kalithea, GR- 176 71, Greece (email: bountos@noa.gr). Ioannis Papoutsis is with IAASARS-NOA and Dimitrios Michail is with IT-HUA. Nantheera Anantrasirichai is with the Department of Computer Science, University of Bristol, Woodland Road Bristol, BS8 1UB, England, UK.}
\author[1]{Ioannis~Papoutsis,~\IEEEmembership{Member,~IEEE}}
\author[2]{Dimitrios~Michail}% <-this % stops a space
\author[3]{Nantheera Anantrasirichai}

\affil[1]{Institute of Astronomy, Astrophysics, Space Applications \& Remote Sensing, National Observatory of Athens}%
\affil[2]{Department of Informatics \& Telematics, Harokopio University of Athens}%
\affil[3]{Department of Computer Science, University of Bristol}
%\thanks{}% <-this % stops a space
%\thanks{.}% <-this % stops a space
%\thanks{Manuscript received April 19, 2005; revised August %26, 2015.}

\maketitle

\begin{abstract}
Ground deformation measured from Interferometric Synthetic Aperture Radar (InSAR) data is considered a sign of volcanic unrest, statistically linked to a volcanic eruption. Recent studies have shown the potential of using Sentinel-1 InSAR data and supervised deep learning (DL) methods for the detection of volcanic deformation signals, towards global volcanic hazard mitigation. However, detection accuracy is compromised from the lack of labelled data and class imbalance. To overcome this, synthetic data are typically used for finetuning DL models pre-trained on the ImageNet dataset. This approach suffers from poor generalisation on real InSAR data. This letter proposes the use of self-supervised contrastive learning to learn quality visual representations hidden in unlabeled InSAR data. Our approach, based on the SimCLR framework, provides a solution that does not require a specialized architecture nor a large labelled or synthetic dataset. We show that our self-supervised pipeline achieves higher accuracy with respect to the state-of-the-art methods, and shows excellent generalisation even for out-of-distribution test data. Finally, we showcase the effectiveness of our approach for detecting the unrest episodes preceding the recent Icelandic Fagradalsfjall volcanic eruption.

\end{abstract}


% \fi
\IEEEpeerreviewmaketitle

\section{Introduction}
%\footnote{
%This work has received funding from the European Union’s Horizon 2020 research and %innovation project DeepCube, under grant agreement number 101004188.
%}
\label{sec:intro}
Globally, 800 million people live within 100 km of a volcano \cite{loughlin2015}. Improvements in forecasting volcanic activity have been shown to reduce fatalities due to volcanic eruptions \cite{auker_statistical_2013}. Hence, several volcano observatories are set-up globally, including the Geohazard Supersites and Natural Laboratories initiative. However, a significant proportion of the $\sim$1,500 holocene volcanoes has no ground-based monitoring, although the deformation at volcanoes is statistically linked to eruption \cite{biggs2014global}, which can be detected prior to the event \cite{FURTNEY201838}. 

Interferometric Synthetic Aperture Radar (InSAR) data from the Sentinel-1, 6-day repeat-cycle, satellite allows the systematic monitoring of volcanic unrest at a global scale. Such abundance of InSAR data have the potential to enable observatories to monitor volcanic activity without additional costs. Fringes detected in wrapped interferograms over volcanoes indicate the onset of deformation, usually due to magma chamber fill-in at depth. Associating fringes with deformation is  non-trivial; atmospheric signals may also give rise to such fringes, which may lead to false positive identifications. Their effect is also amplified in the presence of strong topography, which is usually the case for volcanic domes.

Recent studies have proposed the use of supervised Deep Learning (DL) architectures to automatically detect the presence of ground deformation triggered by volcanic unrest, within single interferograms. Anantrasirichai et~al.~\cite{anantrasirichai2018application} were the first ones to use DL on short-term wrapped interferograms to detect rapid deformation. They rely on heavy data augmentation and employ a transfer learning strategy using AlexNet Convolutional Neural Network (CNN) architecture, pre-trained on ImageNet \cite{imagenet_cvpr09}. Synthetically generated interferograms, based on analytic forward models of magma chambers (e.g. Mogi, dykes, sills), have also been used to train the same network \cite{anantrasirichai2019deep}, reaching 86\% F1-score on real data. Valade~et~al.~\cite{valade2019towards} train a custom CNN architecture on synthetic wrapped interferograms, but test it to only a few real interferograms.
A CNN workflow has also been proposed to identify surface deformation associated with an earthquake \cite{brengman2021identification}, using synthetic wrapped and unwrapped interferograms. They report an accuracy for their best model of 85\%, tested on  32 real InSAR interferograms, yet in principle earthquake induced interferometric signals are denser and clearer than the volcanic ones. Finally, volcanic deformation detection from a single interferogram using a combination of synthetic and augmented real InSAR data has been performed by Gaddes et~al.~\cite{gaddes2021simultaneous}. They apply a VGG-based model, while they split the target class "deformation", to two sub-classes: Sill/Point and Dyke and report an overall accuracy of 83\%.  

%Time-series approaches have also been used to automatically detect deformation. Gaddes et~al.~\cite{gaddes2019using} use Independent Component Analysis and clustering on synthetic unwrapped interferograms to retrieve signals of geophysical significance. Sun et al~\cite{sun2020automatic} address the problem as a denoising, regression one, and use a modified U-Net architecture on synthetic and augmented unwrapped time-series data.  

A common issue encountered by all studies is the scarcity of training data for the positive class, i.e. interferograms with volcanic deformation. Hence, most works cope with class imbalance via heavy data augmentation engineering and the creation of positive class synthetic data for supervised learning. Supervised learning though requires big curated datasets to work well. To counter this, researchers have used classical architectures pretrained on large unrelated datasets. We argue that transfer learning from computer vision tasks provide less meaningful information for the task of volcanic activity detection when compared to features learnt from related data. Indeed the studies discussed above, report that they struggle to generalise well for real, unseen InSAR data.  

Recently, the AI community shifted its focus to resolve these shortcomings of supervised learning, towards unsupervised/self-supervised training schemes. The goal is to exploit the information hidden inside the data and produce features without any human supervision that can generalize well for different classification tasks. 
%Their success has already been proved in natural language processing \cite{devlin2018bert} applications, while Computer Vision applications are catching up too \cite{he2020momentum}.
 In remote sensing some recent works highlight the value of these approaches \cite{9397864}\cite{9345971}.

In this letter, we propose a self-supervised, contrastive learning framework based on SimCLR \cite{chen2020simple} to solve the deformation/non-deformation binary classification problem, using an unbalanced, real, wrapped InSAR dataset. We avoid using unwrapped interferograms or time-series data in order to have faster classification response, which is particularly useful in volcano observatories for near real-time monitoring. Our approach exploits the abundance of unlabelled InSAR data to learn quality visual features, which can be used by a simple linear supervised classifier for the detection task. 
%Given large amounts of unlabelled InSAR data, one can utilize the proposed method for quality feature extraction and then finetune a classifier on a smaller labelled subset.

The contributions of our work are as follows:
\begin{itemize}
    \item We are the first to introduce a self-supervised learning framework for volcanic activity detection.
    \item We propose a training pipeline that does not rely on the generation of massive augmented, synthetic or manually annotated InSAR data.
    \item We demonstrate that training on a large unlabeled InSAR dataset in a self-supervised manner provides more quality features than using pre-trained models from ImageNet. 
    \item Experimental results show that models trained with this framework have the ability to generalize better, even for InSAR data drawn from a different distribution with respect to the training set. 
    \item We provide the first generic feature learning model for InSAR, which can be used for different downstream tasks.
\end{itemize}

\section{Contrastive self-supervised learning pipeline}\label{sec:method}

\begin{figure*}
    \centering
    \includegraphics[width=18cm]{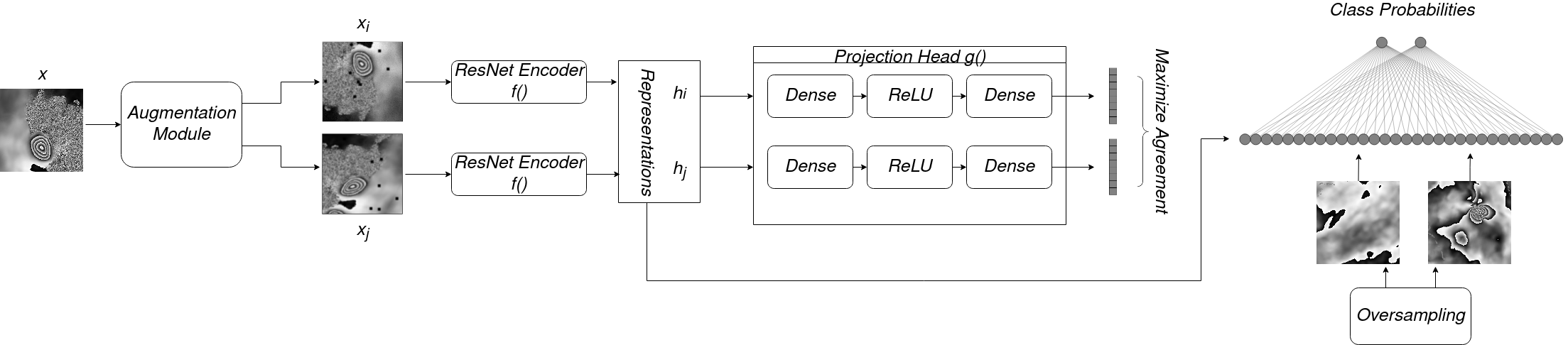}
    % \caption{Interferogram patches are drawn from the training set and fed to the pipeline. For each sample, two augmented views are created. Both are fed to an encoder and then to a projection head to get a vector representation. In this latent space the contrastive loss is applied. After the self-supervised training is complete, we can use the encoder's weights to extract a representation of an input sample. For finetuning, we create balanced batches using oversampling, freeze the encoder's weights and attach a linear classifier on top.}
    \caption{The proposed pipeline. First, we use unlabeled InSAR data to learn feature representations with the SimCLR \cite{chen2020simple} self-supervised framework and then attach a linear classifier for the supervised training with a few labels.}
    \label{fig:simclrpipeline}
\end{figure*}

%Most self-supervised methods typically use one of the following approaches. Invent a handcrafted pretext task e.g Jigsaw puzzles \cite{noroozi2016unsupervised}, hoping that by training a model to solve it, the learnt features would enable it to perform well in the downstream task. The second approach revolves around an instance discrimination task where every image in the dataset belongs to its own class. 

Our approach is set-up as an instance discrimination task where every image in the dataset belongs to its own class. Our pipeline is a two step training process. First, it consists of an encoder trained in a self-supervised manner, and second of a fully connected layer attached on top of the encoder for the supervised classification task. 
For the self-supervised training we utilize the recently introduced SimCLR~\cite{chen2020simple} framework. SimCLR learns representations by trying to maximize the similarity of two augmented views of the same example in the latent space. The main  components of the adopted framework are the following:
\begin{itemize}
    \item A stochastic data augmentation module that creates two different transformations of the same input wrapped interferogram patch. For every patch $x$ the augmentation module creates two augmented views $\tilde{x}_i , \tilde{x}_j$. In our experiments, we use Horizontal and Vertical Flips, Cutout, Multiplicative Noise, Elastic Transformation, Gaussian Blur and Gaussian Noise. Each view is generated from a random combination of these augmentations.
    \item An encoder $f(\cdot)$ for the representation extraction from the augmented patches. There is no constraint on the choice of the encoder. Following \cite{chen2020simple}, we use the ResNet \cite{he2016deep} architecture. The representation is then extracted from the output after the average pooling layer, $h_i = ResNet(\tilde{x}_i)$.
    \item A projection head $g(\cdot)$, that maps the encoder's representation to the space where the contrastive loss is calculated. We, like \cite{chen2020simple} use a multilayer perceptron (MLP) with one hidden layer and a ReLU activation function. Thus, $z_i = g(h_i) = W^{(2)} \sigma(W^{(1)} h_i)$ , where $W^1$ and $W^2$ represent the weight matrices and $\sigma$ is the ReLU function.
    \item The contrastive loss estimation. Given a set $\{ \tilde{x}_k \}$ which contains, among others, a positive pair of augmented interferogram patches $\tilde{x}_i$ and $\tilde{x}_j$,
    the contrastive loss aims to guide our algorithm to identify $\tilde{x}_j$ in $\{\tilde{x}_k\}_{k \neq i}$ for a given $\tilde{x}_i$.
\end{itemize} 

The training pipeline can be seen in Figure~\ref{fig:simclrpipeline} and is as follows. From a batch of size N we create 2N samples using the augmentation process defined above. The augmented InSAR patches created from the same original patch serve as a positive example and the rest 2(N-1) patches in the batch constitute the negative examples. The process can be summarized with the following graph: $x_k \xrightarrow[]{t \in T} \tilde{x}_{i,j}^k \xrightarrow[]{f(\cdot)} h_{i,j}^k \xrightarrow[]{g(\cdot)}z_{i,j}^k$, where T is the set of augmentations. We define the similarity function $sim(x,y)$ as the cosine similarity between $x$ and $y$. For each minibatch we attempt to minimize the following contrastive loss function:
\begin{equation}
    l_{i,j} = -log \frac{e^{sim(z_i,z_j)/\tau}} {\sum_{k=1}^{2N}\mathbbm{1}_{[k \neq i]}e^{sim(z_i,z_k)/\tau}}, 
\end{equation} where $\mathbbm{1}_{[k \neq i]} \in \{0,1\}$ is $1$ if $k \neq i$ and $0$ otherwise and $\tau$ is a temperature parameter. $\tau$ scales the input and expands the range of values of the cosine similarity. We set $\tau=0.5$ and the final loss is calculated on all positive pairs in a mini batch.

Finally, using the encoder that was trained in a self-supervised manner we proceed with the classification task, by freezing the parameters of the encoder and attach a trainable linear classifier on top. We propose an oversampling approach for the supervised classification task, where we randomly choose a class from which the next sample will be drawn. For a batch size N, this process takes place N times. In this approach one sample might be seen more than once in each epoch, thus preventing the domination of the larger class.

\section{Experiments}
All data and code presented in this section are published at the project's repository: (\textit{https://github.com/ngbountos/DeepCubeVolcano}).
\subsection{Datasets}\label{sub:data}

We use only real data that come from two different sources, which we symbolize with S1 and C1 (Table~\ref{tab:datastats}). S1 dataset was provided by the authors of \cite{anantrasirichai2018application} and \cite{anantrasirichai2019deep}, and contains Sentinel-1 wrapped InSAR patches from 16 volcanoes globally. The S1 dataset is highly imbalanced, containing a large number of negative examples but very few positive ones ($\sim$2\%). C1 was manually collected by us from the LiCSAR online InSAR repository \cite{rs12152430} over 5 volcanoes: Taal, Cerro Azul, Fagradalsfjall, Etna and Ale Bagu, and is much more balanced. We use only the S1 dataset for training our models, while we create two different test datasets. The first one contains 64 balanced samples drawn from S1, and the entire C1 dataset serves as a, second, evaluation dataset.

%\begin{table}
   % \centering
    %\begin{tabular}{c|c|c|c}
  %  \toprule
     %   Volcano & \multicolumn{2}{c|}{\#Samples in S1}  & \#Samples in C1\\
        %\hline
      %  - &  Train & Test & Test \\
        %\hline
        
       % Taal & 10 & 3 & 11\\
        %San Pablo Volcanic Field & 2 & 0& 0\\
        %Rinjani & 13 & 0 & 0 \\
        %Laguna del Maule & 18 & 9 & 0\\
        %Sierra Negra & 64 & 6 & 0\\
        %Cerro Azul & 11 & 1 & 212\\
        %Reykjanes & 2 & 0 & 0 \\
        %Domuyo & 1 & 1 & 0\\
        %Hayli Gubbi & 1 & 0 & 0\\
        %Fentale & 1 & 1 & 0\\
        %Erta Ale & 12 & 4 & 0\\
        %Corbetti & 2 &0 & 0\\
        %Bora Ale &13 & 2 & 0 \\
        %Fagradalsfjall & 0& 0 & 125 \\
        %Etna & 0 & 4 & 54\\
        %Ale Bagu & 0& 1 & 2\\
        %\bottomrule

 %   \end{tabular}
%\caption{The volcanoes for which we used InSAR patches.}
%    \label{tab:deform}
%\end{table}

\begin{table}
    \caption{Overview of the datasets used in this work. For the self-supervised task, the positives and negatives of S1 are employed together without labels. C1 is used only for evaluation.}\label{tab:datastats}
    \begin{tabular}{ccc|ccc}
    \toprule
   %\hline
    %&\multicolumn{1}{c|}{}& \\% \cline{2-5}
    Data Source &
    \multicolumn{2}{c}{Train}& \multicolumn{2}{c}{Test} &Total\\ \hline
     - & Positive & Negative& Positive& Negative  & \\ \hline
    S1&150&7386&32&32&7600\\ \hline 
    C1 & -&-&404&365&769\\ \hline \\
    \end{tabular}

\end{table}

\begin{figure}
    \centering
    \includegraphics[width=\columnwidth]{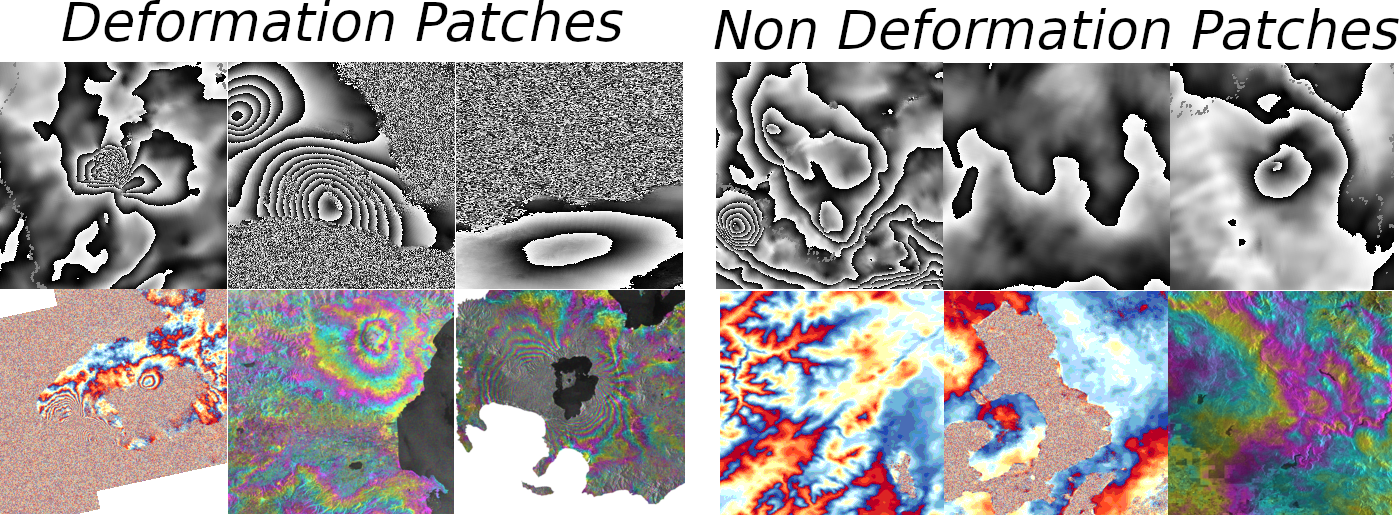}
    \caption{Sample interferogram patches from both data sources. The first and second rows contain samples from S1 and C1 sources respectively. C1 dataset is diverse - from left-to-right and for the deformation class: 1) Cerro Azul unfiltered, interferometric phase only, 2) Etna descending interferogram, Goldstein filtered, phased and amplitude overlay, and 3) Taal, ascending interferogram, water masked interferometric phase, Goldstein filtered.}
    \label{fig:datasetgrid}
\end{figure}

The two test sets S1 and C1 are quite different. Figure~\ref{fig:datasetgrid} shows examples of deformation and non-deformation samples from both sources. C1 test dataset is highly diverse as we have included 1) wrapped interferograms from both descending and ascending viewing geometries, 2) unfiltered and Goldstein phase-filtered interferograms, and 3) pure phase-only interferograms and interferograms where the phase is overlaid with SAR amplitude. S1 dataset is much more harmonised and therefore S1 and C1 are considered to be drawn from distributions with different characteristics with respect to noise level and visual features. Such challenging test samples are used to evaluate the generalization performance of our self-supervised approach. 

%The two test sets from S1 and C1 have been produced by different InSAR processing pipelines and therefore we assume that the test samples are drawn from different distributions [characteristics?]. Figure~\ref{fig:datasetgrid} shows examples of deformation and non-deformation samples from both sources. The two data sources are quite different. C1 test dataset is highly diverse as we have included 1) wrapped interferograms from both descending and ascending viewing geometries, 2) unfiltered and Goldstein phase-filtered interferograms, and 3) pure phase-only interferograms and interferograms where the phase is overlaid with SAR amplitude. Such challenging test samples will be used to evaluate the generalization performance of our self-supervised approach. 

\subsection{Contrastive learning model performance}\label{sub:features}
%Quality of trained features and comparison with state-of-the-art methods
We evaluate the performance of the encoder trained with our proposed self-supervised approach (Section~\ref{sec:method})
and compare it to the pretrained features from ImageNet~\cite{imagenet_cvpr09}. We experiment with different scales of the ResNet architecture and 
utilize the linear evaluation protocol, i.e. we freeze the encoder parameters and fine-tune a simple linear classifier on top of the network. The linear evaluation protocol is a standard way to assess the quality of the learnt representations \cite{he2020momentum}.
To speedup convergence for the self-supervised stage, we initialize the weights of the encoder with the parameters learnt from ImageNet. We then retrain all layers with the self-supervised method for 200 epochs using an unlabelled version of the S1 dataset. At the fine-tuning step we apply the oversampling approach (Section~\ref{sec:method}) to the labeled S1 dataset. No oversampling is performed for the self-supervised stage, as it assumes no class knowledge.

%In our experiments we evaluate the performance of different models, in both pretrained schemes. From now on we will refer to the ImageNet based pretraining scheme as setting 1 and the proposed (see section \ref{sec:method}) method as setting 2. The models we examine are: ResNet18, ResNet50, ResNet101, VGG16 \cite{simonyan2014very} and AlexNet \cite{krizhevsky2012imagenet}. At first we provide a comparison between the performance of the ResNet architectures in setting 1 and 2. We then compare the best of these models with networks used in current state-of-the-art i.e AlexNet and VGG16. The goal of our experiments is to assess the quality of the features learnt at the pretraining stage. To do that, we utilize the linear evaluation protocol i.e we freeze the encoders parameters and finetune a linear classifier on top of the network.
%Αt the finetuning step we use the oversampling approach mention on section \ref{sec:method}. This is not the case for the pretraining step.

Table~\ref{tab:comp} shows how ResNet architectures with different capacities compare in the two different test setups. The models we 
examine are ResNet18, ResNet34 and ResNet50. 
It is noteworthy that the results using the features learnt from the contrastive learning framework were obtained after only 1 to 3 epochs of fine-tuning the linear classifier. %To be more precise, we finetune ResNet50 for 1 epoch, ResNet34 for 3 and ResNet18 for 2.
We found that setting the learning rate between  $0.001$ and 0.005 works best. 
% (If 300 epochs works well )As noted earlier, batch size is an important hyperparameter for the SimCLR framework, as larger batch sizes will provide more negative samples. In our experiments we are restricted by our computational resources. We thus use batch size of 112 for ResNet18, 32 for ResNet50 and 16 for ResNet101. To counter the small batch size for the ResNet101 we pretrain for more epochs. For the rest of the models we set the pretraining duration to 200 epochs.
On the contrary, the networks that used the pre-trained encoder from ImageNet required 75 epochs to converge. We set the learning rate to 0.001. For the fine-tuning step we use a batch size of 112. At the pre-training stage, we use the largest possible batch size depending on the architecture. We set the batch size to 32 for ResNet50 and 112 for ResNet18 and ResNet34. %ResNet101 Simclr lr = 0.005

 We conduct two additional experiments using ResNet50, our best performing backbone encoder. First, we train SimCLR from scratch using random initialization for 300 epochs to show that the performance gain comes from self-supervised training on InSAR data alone. We publish this model on the project's repository. Second, we test MoCo~\cite{he2020momentum}, a different self-supervised approach to show that performance gain can be achieved from different self-supervised methods as well.

\begin{table*}[ht]
\centering
\caption{Experiment results on both test sets. ACC, FP, TP, FN, TN, F1, P and R, stand for overall accuracy, false positives, true positives, false negatives, true negatives, f1-score, Precision and Recall, respectively. }\label{tab:comp}
\begin{tabular}[t]{ccccccccc|cccccccc}
\toprule
    & &  &  & & S1& & & &  & & & & C1& & &\\
    \hline
    {Model} & {ACC} & {FP} & {TP} & {FN} & {TN}& F1 & P & R & {ACC} & {FP} & {TP} & {FN} & {TN} & F1 & P & R \\ 
    \hline
    ResNet18-ImageNet & 81\% & 12 & 32 & 0 & 20 & 0.841&0.727 &1 &  64\% & 0 & 132 & 272 & 365 & 0.491 &1 &0.326 \\ 
    ResNet18-SimCLR & 84\textbf{\%} & 9 & 31 & 1 & 23  & 0.860 &0.775  &0.968 & 70\% & 0 & 178 & 226 & 365  & 0.611 & 1 & 0.440\\  
    ResNet34-ImageNet & 82\% & 10 & 31 & 1 & 22& 0.848& 0.756& 0.968 & 70\% & 3 & 181 & 223& 362& 0.615 & 0.983 &0.448\\
    ResNet34-SimCLR &82\%& 11 & 32 & 0  & 21 &0.853& 0.744 & 1 &  \textbf{91\%} & 4 & 339& 65 & 361 & 0,907 & 0.988 & 0.839 \\
    ResNet50-ImageNet  & 82\% & 10 & 31 & 1 & 22 & 0.848 &0.756 & 0.968 &  63\% & 1 & 125 & 279 & 364 &0.471 &0.992 &0.309 \\
    ResNet50-SimCLR & 85\% & 8 & 31 & 1 & 24& 0.872 & 0.794& 0.968& \textbf{91\%} & 10 & 347 & 57 & 355 &\textbf{0.911 }&0.971 & 0.858 \\
    ResNet50-SimCLR-Scratch& 85\% &8 & 31 & 1& 24& 0.872& 0.794 & 0.968 &86\% & 2& 306& 98 & 363 &0.859 & 0.993& 0.757\\
    ResNet50-Moco & 79\% & 13 &32 &0&19& 0,831& 0.711&1 &  82\% & 0 & 267 & 137 & 365 & 0.795& 1 & 0.660\\
    \midrule
    AlexNet & 82\% & 9 & 30 & 2 & 23& 0.844 & 0.769& 0.937 & 72\% & 49 & 244 & 160 & 316 & 0.699 &0.832 &0.603\\
    VGG16 & 85\textbf{\%} & 6 & 29 & 3 & 26 & 0.865 &0.828 &0.906 & 64\% & 39 & 171 & 233 & 326 &0.556 & 0.814 &0.423 \\
    \midrule
    ViT-ImageNet & 90\% & 3 & 29 & 3 & 29 & 0.906 & 0.906 & 0.906 & 59\% & 5 & 96 & 308 & 360&  0.343 & 0.950 & 0.343\\
    DenseNet121-ImageNet & 82\% & 9 & 30 & 2 & 23 & 0.844 & 0.769 & 0.937 & 54\% &0 & 53 & 351& 365 & 0.231 & 1 & 0.131 \\
    InceptionV4-ImageNet & \textbf{92}\% & 4 & 31& 1 &28 &\textbf{ 0.924} & 0.885& 0.968& 69\% & 23& 196& 208 & 342 & 0.628 & 0.894 & 0.485\\
    \bottomrule \\
\end{tabular}
\end{table*}

Additionally, we compare our models with the state-of-the-art - all published methods use supervised approaches. 
We train AlexNet's and VGG16's final layer for 50 epochs, while keeping the rest of the layers freezed, with the ImageNet pre-trained weights.
We also examine three more methods, popular in computer vision i.e Vision Transformer (ViT), DenseNet121 and Inception-V4 (Table~\ref{tab:comp}). We use oversampling and random rotation augmentation for all methods.%

\subsection{Discussion}
\label{sub:discussion}
The results summarised at Table~\ref{tab:comp} show that the models trained with the SimCLR method performed better or comparable with the respective ones pre-trained with ImageNet, for each test dataset and for each ResNet encoder architecture. This is significant; it highlights the fact that training in a self-supervised approach with 7,536 unlabeled samples only (Table~\ref{tab:datastats}), drawn from a distribution of wrapped interferograms, provides better quality features than using models pre-trained in a supervised way from $\sim$1.5 million ImageNet images. It is even more impressive that our proposed self-supervised learning technique  produced models that required only 1-3 epochs of finetuning to achieve these results, versus the 75 epochs needed for the Imagenet pre-trained model, for the same task.

The major enhancement in our approach however, shows itself at the C1 dataset. The high quality of the learnt InSAR data representations is clearer in the second half of Table~\ref{tab:comp} that summarises the experiments on the C1 test dataset, which is drawn from a distribution with different characteristics with respect to the training set (Section~\ref{sub:data}). While for S1 dataset the best ImageNet and SimCLR models provide 92\% and 85\% overall accuracy respectively, for C1 dataset the corresponding accuracies are 70\% and 91\%. The supervised ImageNet model struggles to resolve the required features from the new InSAR data distribution and overall, this large performance gap underlines the ability of the self-supervised model to generalize better. To further validate this we construct 731 synthetic interferograms using SyInterferoPy~\cite{gaddes2021simultaneous} over a collection  of  subaerial  volcanoes. Deformation patterns include those due to dykes, sills, and Mogi sources. Our ResNet50-SimCLR model reaches 88.9\%, while ResNet50-ImageNet 69.9\% and InceptionV4-ImageNet 66.2\% true positive rate respectively. Again the self-supervised learnt features generalize  better.

The last five rows of Table~\ref{tab:comp} provide a comparison between ResNet50-SimCLR, our best performing encoder, and architectures on the same task, proposed by the state-of-the-art studies of Section~\ref{sec:intro} and other popular methods used in computer vision. Again, the proposed method generalizes better, as seen in the C1 part of the table. On S1 all models perform well with Inception-V4 achieving the best accuracy. On C1, our method reaches 91\% overall accuracy and 0.911 F1 score based on the linear evaluation protocol. 
%To the best of our knowledge these are the highest reported scores achieved for the task of deformation detection from single wrapped interferograms, although other studies have used different datasets. 

In addition, our pipeline did not make use of massive augmented, or synthetically generated datasets, as opposed to the state-of-the-art approaches (Section~\ref{sec:intro}). Equally important, the achieved performance has been reached with a training set containing just 150 manually annotated InSAR patches with deformations. The potential from exploiting large unlabelled InSAR datasets in a self-supervised approach is at least promising. Given the recent availability of online repositories, such as LiCSAR used in this work, that contain and produce hundreds of interferograms over volcanoes globally, the deployment of this pipeline on a large distributed system, increasing the effective batch size is a natural next step. %Extracting quality features from patches containing volcanoes from all over the world will lead to high quality models that can identify ground deformations and alarm observatories globally, in time, using only a small subset for finetuning.
In fact, greater computational resources lead to better models. 
Using larger batch sizes, training longer and optimizing the stochastic augmentations used for SimCLR, improves the performance of contrastive learning \cite{chen2020simple}. Furthermore, our analysis showed that Elastic Transformation is especially important for data augmentation, potentially due to the nature of fringe patterns within InSAR data, either these patterns are related to deformation, atmospheric disturbances, topographical errors, or orbital ramps, etc.
In our work we also use the maximum batch size possible depending on the model's architecture and available memory.
%This is also validated with the class activation maps presented in Figure~\ref{grid:activations}.

\section{Fagradalsfjall volcanic eruption case study}
\label{sec:case}
We apply our approach on Fagradalsfjall volcano at Reykjanes Peninsula, Iceland. We focus on two recent unrest episodes. Triggered by dyke intrusions, the inflation episodes took place in mid-January 2020, and early March 2021. The latter episode led to an effusive eruption, on 19 March 2021, -- the first known eruption on the peninsula in about 800 years.  

\begin{figure*}[ht]
%\begin{tabular}{ccc}
%images from Resnet50 trained on Merged finetuned on merged dataset 
 %\includegraphics[width=.3\linewidth]{images/iceland1.png} %& \includegraphics[width=.3\linewidth]{images/iceland2.png%} & \includegraphics[width=.3\linewidth]{images/iceland3.p%ng}
 \includegraphics[width=18cm]{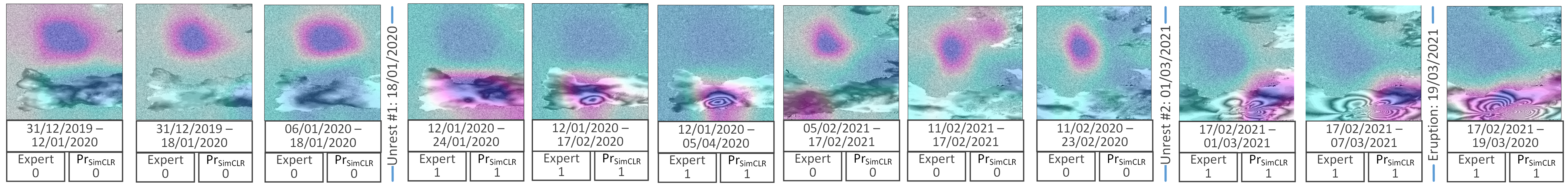}
%\end{tabular}
\caption{Visualization of ResNet50 activations on Fagradalsfjall volcano. Pink represents the area that affected the network's decision the most. Both unrest episodes are shown in chronological order. $Pr_{SimCLR}$ and Expert are the predictions made by our method and the InSAR expert, respectively (1=positive, 0=negative).}
\label{grid:activations}
\end{figure*}
In order to simulate a real, working product, we merge all our datasets and collect some extra InSAR patches to finetune our models and produce a quality classifier. We unfreeze the fully connected layer as well as layer 3 and 4 of the ResNet50 SimCLR encoder. We train for 2 epochs and reduce the learning rate to $0.0005$. In total the new training set contains 614 deformation and 7872 non deformation patches. We feed the network with two time-series of wrapped interferograms, one for each unrest event.

Figure~\ref{grid:activations} presents the model classification decision for each single interferogram vis-à-vis the decision from an InSAR expert. In addition, the figure shows the areas of the patch that affect the most the decisions of the final, fully connected layer. This visualization was produced using the Class Activation Mapping (CAM) technique~\cite{zhou2016learning}. There are two key observations drawn from this use case. First, the model trained with the proposed method focuses on the correct patterns (the fringes) of the interferogram patch. Second, the pipeline correctly captures the start of both unrest episodes, triggering a potential alarm.

\section{Conclusion}

In this work, we implemented a pipeline to train binary classification models for volcanic unrest detection utilizing unlabelled InSAR datasets, thus without the the need to create huge labelled datasets or generate error-prone synthetic data. We provided proof for the superiority of the self-supervised learnt features when compared to models pre-trained from ImageNet, and the ability of our approach to generalise effectively even for out-of-distribution test samples. Our approach outperforms state-of-the-art supervised methods.

Volcanic unrest early warning is of major importance for civil protection authorities and volcano observatories. Setting-up alert mechanisms enhances response effectiveness and allows for scientists to deploy critical in-situ monitoring equipment to assess more accurately volcanic hazard. We highlighted that this could be implemented in the case of the 2020-2021 Fagradalsfjall volcano unrest and eruption.

Finally, we believe that there is much potential in our self-supervised approach, given the abundance of InSAR data produced regularly by the Sentinel-1 missions. Exploiting the information they contain in a self-supervised way, while labelling only a small subset paves the way towards a global volcanic unrest detection system, but may also be applicable to a plethora of other remote sensing applications and tasks.

\ifCLASSOPTIONcaptionsoff
  \newpage
\fi

%\clearpage



%\bibliographystyle{IEEEtran}

\bibliographystyle{ieeetr}
\bibliography{arxiv.bib}

%\begin{thebibliography}{1}

%\end{thebibliography}


%\end{IEEEbiography}


%\newpage



%\vfill


\end{document}